\documentclass[10pt,twocolumn,letterpaper]{article}

\usepackage[pagenumbers]{cvpr} 

\usepackage[dvipsnames]{xcolor}

\usepackage{xcolor}
\usepackage{xspace}
\usepackage{wrapfig}
\definecolor{light-gray}{gray}{0.95}

\newcommand{\reals}{\mathbb{R}}

\usepackage{duckuments}

\usepackage{overpic}
\usepackage[usestackEOL]{stackengine}
\usepackage{varwidth}
\usepackage{setspace}

\definecolor{cvprblue}{rgb}{0.21,0.49,0.74}
\usepackage[pagebackref,breaklinks,colorlinks,citecolor=cvprblue]{hyperref}

\usepackage{xspace}
\newcommand{\name}{\textsc{LooseControl}\xspace}

\title{\name: Lifting ControlNet for Generalized Depth Conditioning }

\author{Shariq Farooq Bhat\\
KAUST
\and
Niloy J. Mitra\\
University College London, Adobe Research\\
\and
Peter Wonka\\
KAUST\\
}

\begin{document}
\twocolumn[{%
\renewcommand\twocolumn[1][]{#1}%
\maketitle
\vspace*{-.35in}
\begin{center}
     \includegraphics[width=\linewidth]{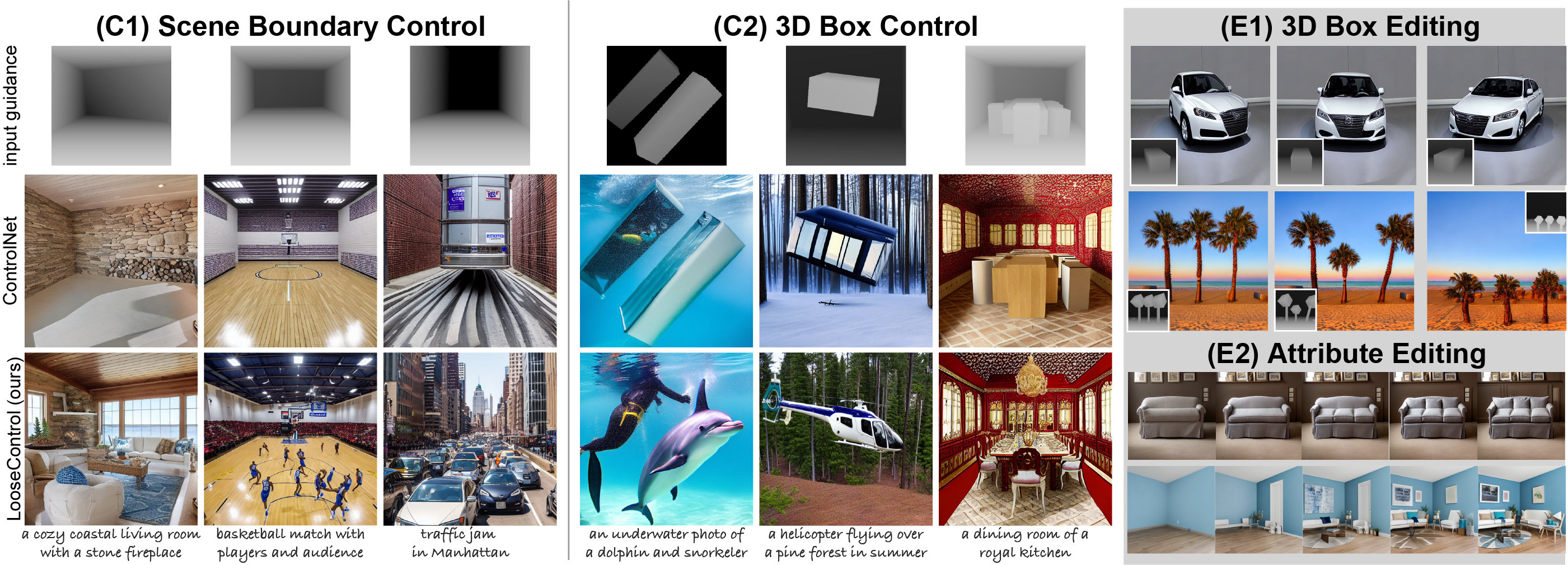}
    \captionof{figure}{\captionsize Our framework LooseControl enables multiple ways to control the generative image modeling process. \textit{Left:} (C1) Scene boundary control lets the user specify the boundary of the scene. We show the control inputs on top, the ControlNet results in the middle, and our results at the bottom. \textit{Middle:} (C2) 3D box control can additionally specify object locations with the help of approximate 3D bounding boxes. We again show the control inputs on top, the ControlNet results in the middle, and our results on the bottom. \textit{Right-Top:} (E1) 3D box editing. Note how the overall style of the scene is preserved across edits. \textit{Right-Bottom:} (E2) Attribute editing: an example of changing a couch to another one and changing the overall furniture density in a room.} 
    \label{fig:teaser}
\end{center}
}]

\begin{abstract}
We present \name to allow generalized depth conditioning for diffusion-based image generation. ControlNet, the SOTA for depth-conditioned image generation, produces remarkable results but relies on having access to detailed depth maps for guidance. Creating such exact depth maps, in many scenarios, is challenging. This paper introduces a generalized version of depth conditioning that enables many new content-creation workflows. Specifically, we allow (C1)~scene boundary control for loosely specifying scenes with only boundary conditions, and (C2)~3D box control for specifying layout locations of the target objects rather than the exact shape and appearance of the objects. Using \name, along with text guidance, users can create complex environments (e.g., rooms, street views, etc.) by specifying only scene boundaries and locations of primary objects. Further, we provide two editing mechanisms to refine the results: (E1)~3D box editing enables the user to refine images by changing, adding, or removing boxes while freezing the style of the image. This yields minimal changes apart from changes induced by the edited boxes. (E2)~Attribute editing proposes possible editing directions to change one particular aspect of the scene, such as the overall object density or a particular object. Extensive tests and comparisons with baselines demonstrate the generality of our method. We believe that \name can become an important design tool for easily creating complex environments and be extended to other forms of guidance channels.
Code and more information is available at \url{https://shariqfarooq123.github.io/loose-control/}.\footnote{The project was supported in part by ``NTGC-AI" funding at KAUST.}
\end{abstract}

\section{Introduction}
\label{sec:intro}

Diffusion-based generative models now produce images with a remarkable degree of photorealism. ControlNet~\cite{zhang2023adding},  trained on top of StableDiffusion~\cite{rombach2022high}, is the most powerful way to control such a generation process. Specifically, ControlNet allows guidance in the form of one or more of depth, edges, normal, or semantic channels. This ability to accurately control the image generation process enables various creative applications without requiring different architectures for different applications. However, providing perfect guidance for ControlNet can itself be challenging. 

For example, imagine a scenario where a user wants to create living room images using depth guidance. She is now expected to supply a depth map containing information about walls, room furniture, all the way to even smaller objects and decorations. 
This is non-trivial. Producing realistic depth maps, especially for cluttered scenes, is probably as challenging as solving her original task. On the one hand, providing rough guidance, e.g., providing depth information for walls, floor, and ceiling, results in unsatisfactory images. While one may expect a furnished room, the room will be empty. Also providing approximate depth using target bounding boxes for furniture does not yield the desired result, because only boxy objects are generated. Figure~\ref{fig:teaser} shows some examples using ControlNet~\cite{zhang2023adding}. 

We present \name that allows controlling image generation using generalized guidance.  In this work, we focus on guidance through depth maps.
We consider two types of specifications. (C1)~\textbf{Scene boundary control} where a layout can be given by its boundaries, e.g., walls and floors, but the final scene can be filled by an arbitrary number of objects that are not part of the depth conditioning as long as these additional objects are closer to the camera than the scene boundary. (C2)~\textbf{3D box control} where, in addition to layout boundaries, users can provide finer-scale guidance in the form of approximate bounding boxes for target objects. In the final generation, however, there can be additional secondary objects that do not need to strictly adhere to the given boxes.
We demonstrate that many layout-guided image generation scenarios (e.g., rooms, streets, underwater) can be addressed in this framework. Figure~\ref{fig:teaser} (left, middle) shows some of our generations. 

We also provide two interactive editing modes to refine the results.
(E1)~\textbf{3D box editing} enables the user to change, add, and remove boxes while freezing the style of the resulting image. The goal is to obtain minimal changes apart from the changes induced by the edited boxes. We describe how a notion of style can be formalized and preserved in our diffusion setup. 
(E2)~\textbf{3D attribute editing} enables users to change one particular attribute of the result, such as the type of one particular piece of furniture. Since diffusion processes return a distribution of results, we perform local analysis around any generation to reveal dominant modes of variation. 
Figure~\ref{fig:teaser} (right) shows a few editing examples. 

The technical realization of these four components is built on ControlNet with a frozen StableDiffusion backbone.
We first propose a Low Rank~(LoRA) based network adaptation.
The LoRA-based architecture allows the network to be fine-tuned in a few steps with only a small amount of training data, preventing the network from forgetting the original generation weights.
A second important component is automatically synthesizing the necessary training data without requiring manual annotations.
Finally, the edits are realized by manipulating the ``keys" and ``values" in attention layers, and a singular value analysis of the ControlNet Jacobian.

We evaluate a range of use cases. Since we enable a strictly more general depth control, there are no direct competitors. Hence, we compare strict and weaker guidance by varying the control weighting in ControlNet, as well as retraining ControlNet on synthetic data. We evaluate our setup on a variety of different scenes and also perform a user study that reveals over $95\%$ preference for our method compared to baselines. In summary, we are the first to allow image generation from loose depth specifications and provide multiple types of edits to semantically explore variations in generations.

\section{Related Work}
Diffusion models have rapidly evolved as a leading generative approach, demonstrating remarkable success both in 2D image generation~\cite{ho2020denoising,nichol2021improved, dhariwal2021diffusion, rombach2022high, nichol2021glide, ramesh2022hierarchical, balaji2022eDiff-I,saharia2022photorealistic} as well as 3D shape generation~\cite{hui_neural_2022,shue_3d_2023,chou_diffusion-sdf_2023,zheng_locally_2023,zhang_3dshape2vecset_2023,erkocc2023hyperdiffusion,karnewar2023holofusion,single-stage-diffusion}. While there are many similarities and synergies between 2D and 3D diffusion, in this paper, we only focus on the problem of adding conditional control to text-to-image diffusion models. In this context, many existing methods proposed solutions on direct guidance through well-known condition inputs like inpainting masks~\cite{xie2023smartbrush, wang2023imagen}, sketches~\cite{voynov2023sketch}, scene graphs~\cite{yang2022diffusion}, color palletes~\cite{vavilala2023applying,mou2023t2i,huang2023composer}, 2D bounding boxes~\cite{li2023gligen}, segmentation maps~\cite{rombach2022high, wang2022pretraining, couairon2022diffedit}, composition of multiple text descriptions~\cite{liu2022compositional}, depth maps~\cite{zhang2023adding,mou2023t2i} or by fine-tuning these models on a few subject-specific images~\cite{gal2022image,mokady2023null,ruiz2023dreambooth,Tewel2023KeyLockedRO}. We focus on introducing new types of control.
ControlNet~\cite{zhang2023adding} stands out as a key method in spatial control, supporting a diverse array of conditions like edge maps, depth maps, segmentation masks, normal maps, and OpenPose~\cite{cao2019personopenpose, simon2017handopenpose, cao2017realtimepersonopenpose, wei2016cpmPosemachines} under a single framework. It is notably based on the widely-used open-source model, Stable Diffusion~\cite{rombach2022high}, which contributes to its popularity. However, creating spatial control conditions manually by a user for ControlNet can be challenging, often resulting in indirect control where conditions are derived from another image source, which can stifle the creative generation process. Additionally, this method can be restrictive in terms of the diversity of generations per condition, posing limitations in various scenarios.
Our work builds on top of ControlNet and introduces novel conditioning mechanisms that are not only easy to construct manually by a user but also offer enhanced control and flexibility. In particular, we contribute novel forms of loose control described in the next section.

\begin{figure*}[t!]
    \centering
    \includegraphics{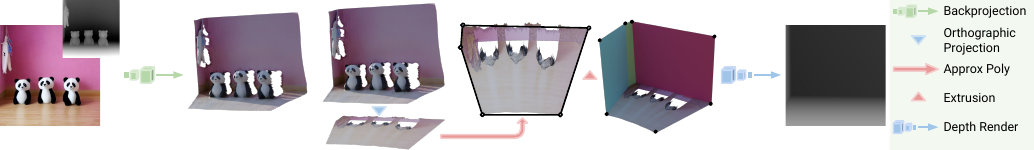}
    \caption{Pipeline for extracting boundary depth from an image. From left to right: Input image and its estimated depth map, back-projected 3D mesh, orthographic projection of the mesh on a horizontal plane, polygon approximation of the 2D boundary, extrusion of poly sides, resulting boundary depth map. For ease of visualization, the ceiling is not shown.}
    \label{fig:pipeline-boundary-depth}
\end{figure*}

\section{Problem Setup - \name}
We formally introduce the core problem statement proposed in this work - \textit{loose depth control}. 
Ordinary depth control, as implemented by the original ControlNet can be formally described as follows: Given an input condition depth map ${D}_{c}$, and access to an off-the-shelf monocular depth estimator $f_D$, generate an image $I_{gen}$ such that the estimated depth $f_D(I_{gen})$ respects the input depth condition \ie:
\begin{equation}
\label{eq:ordinary-control}
    \text{Generate $I_{gen}$ such that } f_D({I}_{gen}) = D_{c}.
\end{equation}

Thus, by design, the conditioning imposed by ControlNet is strict: as per training, the equality in \cref{eq:ordinary-control} must \textit{exactly} hold. Our goal is to extend this notion of control to a more flexible generalized form. We therefore define \textit{generalized control} via the following setup: Given an input condition depth map ${D}_{c}$, and access to an off-the-shelf monocular depth estimator $f_D$, and an arbitrary Boolean condition function $\phi(\cdot, \cdot)$:
\begin{equation}
\label{eq:generalized-control}
    \text{Generate $I_{gen}$ such that } \phi(f_D(I_{gen}), D_{c})\text{ is \textsc{true}}.
\end{equation}

It is easy to observe that \cref{eq:ordinary-control} is a special case of \cref{eq:generalized-control}. In this work, we propose to consider two other cases: scene boundary control and 3D box control, as described next.

\textbf{(C1)~Scene boundary control.}
In this case, we impose the condition such that the input depth condition $D_c$ only specifies the tight upper bound of depth at each pixel:
\begin{equation}
\label{eq:boundary-control}
    \phi : f_D(I_{gen}) \leq D_c.
\end{equation}

\textbf{(C2)~3D box control.}
In this case, we let the condition $D_c$ control only the approximate position, orientation, and size of objects by specifying their approximate 3D bounding boxes. This leads to finer control than (C1) yet still avoids specifying the exact shape and appearance of the objects. Essentially, we design a condition function $\phi$ that ensures that the objects $O_{gen}^i$ generated in an image conform to their respective specified 3D bounding boxes $B_i$:
\begin{equation}
\label{eq:box-control}
    \phi : B_i\sim\text{3DBox}(O_{gen}^i) \quad\forall i,
\end{equation}
where $\text{3DBox}(O_{gen}^i)$ represents the oriented 3D bounding box of the $i$-th object segment in the generated image. This means that the position, size, and orientation of each object will be approximately within the bounds set by its corresponding 3D bounding box. Although our 3D box control training is strict, we show that boxes can be treated as only approximate, which proves to be highly beneficial in practice.

Both \cref{eq:boundary-control} and \cref{eq:box-control} specify a form of depth condition that is strictly more general than the ordinary control such as realized by ControlNet. We term such conditioning as \name. 

\section{Realizing \name}
To realize \textit{ordinary} depth control of text-to-image diffusion models, one needs access to triplets of the form $(T, D_c, I)$ for training, where $T$ is the text description of the image $I$ and $D_c$ represents the depth condition. Generally, one has access to the pairs $(T,I)$ and the ordinary depth condition $D_c$ is obtained by applying an off-the-shelf depth estimator $f_D$ on the given image (i.e., $D_c = f_D(I)$) to obtain the triplets $(T, D_c=f_D(I), I)$. However, in our case, the depth condition $D_c$ must take a more generalized form. For our goal, the form of $D_c$ is constrained by three main requirements: \textbf{(i)} Compatibility with ControlNet, where the depth condition should resemble a conventional depth 
map. \textbf{(ii)} Efficient to extract from a given image without manual annotation. \textbf{(iii)} Easy to construct manually by a user. We describe below how to obtain the appropriate form of $D_c$ such that \name is realized.

\subsection{How to represent scenes as boundaries?}
\label{subsec:how-to-boundary}

\begin{figure}[h!]
    \centering
    \includegraphics[width=\columnwidth]{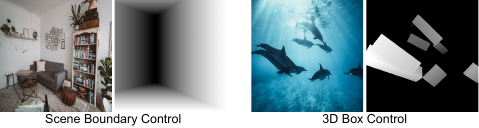}
    \caption{Illustration of proxy depth for scene boundary control training (left) and 3D box control training (right).}
    \label{fig:proxy-depth-illustration}
    \vspace{-1.5em}
\end{figure}

We begin by outlining the estimation of the depth condition $D_c$ from a given image for implementing scene boundary control as described in \cref{eq:boundary-control}. In this context, we seek a depth condition that acts as an upper depth limit for individual pixels. We propose to extract the scene boundary surface for upper-bound training in the following manner. Specifically, we define the boundary as a set of planar surfaces encompassing the scene that accurately delineates the scene's spatial extent. For a given bounded 3D scene and a specific camera viewpoint, the boundary depth is defined as the z-depth of these boundary surfaces. In practice, this means that all pixels in the image inherit the depth value associated with the boundary surface, even if the boundary surfaces are occluded by objects within the scene's interior. To provide a concrete example, consider an indoor scene (refer to \cref{fig:proxy-depth-illustration}). The boundary typically encompasses the walls, ceiling, and floor planes. Consequently, a boundary depth map for an indoor scene exclusively reflects the depths of walls, ceiling, and floor, irrespective of the presence of room furniture. 

A naive approach to extracting boundary depth involves leveraging annotated room layout datasets. However, such datasets often contain ambiguous annotations in multi-room images, and the room-level annotations directly violate our ``upper" bound condition (see Supplementary Materials). Additionally, this approach necessitates manual annotation for boundary extraction and confines the scope to room-centric scenarios. Our objective is to devise a strategy that is applicable across diverse scene types and readily applicable to extract boundary depth from any given image. To this end, we propose the following algorithm: 

We use a multi-step approach to efficiently extract boundary depth from a given image. Refer to \cref{fig:pipeline-boundary-depth} for the outline of our approach. We begin by estimating the depth map of the given image using an off-the-shelf monocular depth estimator~\cite{bhat2023zoedepth}. We then back-project the image into a 3D triangular mesh within the world space. Our goal is to extract the planar surfaces that encompass this mesh. For efficiency, we only make use of vertical planes during training. This reduces the 3D boundary extraction problem to 2D. We project the 3D mesh of the scene onto a horizontal plane using orthographic projection. This projection facilitates the precise delineation of the 2D boundary that encapsulates the scene. We then approximate the 2D boundary of projection with a polygon and extrude the sides into planes matching the scene height. For an indoor scene, for example, these `poly'-planes represent a good approximation of wall planes. These poly planes and optionally the ground and ceiling plane (if available) together form the boundary surface. We then render the depth of this boundary surface from the original camera view to get the boundary depth map $D_c$, serving as a proxy for the scene boundary condition.

\subsection{How to represent scenes as 3D boxes?}

We now describe the estimation of $D_c$ from a given image to realize \textit{3D box control} (\cref{eq:box-control}). For a given image, our goal is to construct $D_c$ such that it contains the information about the 3D bounding boxes of the objects present in the scene. At the same time, we need to ensure compatibility with depth-conditioned frameworks, such as ControlNet, where the depth condition should resemble a conventional depth map. To achieve this, our idea is to obtain 3D bounding boxes of the objects in the image and render the depth of the boxes. The resulting depth map can serve as a proxy for specifying 3D bounding boxes (See \cref{fig:proxy-depth-illustration}).
However, this task necessitates a robust monocular 3D object detector, which, currently, presents challenges. Existing monocular 3D object detection pipelines tend to be either sparse, targeting specific object categories, or domain-specific, focusing on scenarios like road scenes, limiting their versatility. To address these limitations, we introduce a custom monocular pipeline that approximates the 3D bounding boxes of objects and is simultaneously dense (capable of recovering any object type) and generalizable (no domain preference). 

\begin{figure*}[t!]
    \centering
    \includegraphics[width=\linewidth]{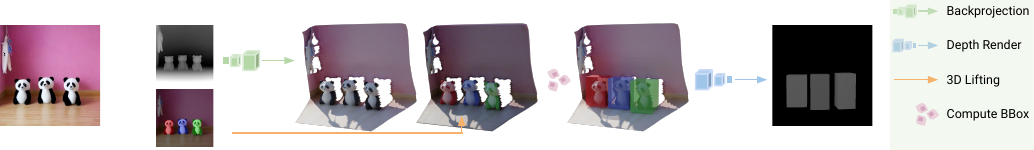}
    \caption{Pipeline for extracting proxy depth for 3D box control from an image. Left to right: Input image, its estimated depth and segmentation maps, back-projected 3D mesh, 3D lifting of segmentation, 3D bounding boxes for segments, resulting proxy depth map.}
    \label{fig:pipeline-bbox-det}
\end{figure*}

For a given image, we first obtain its depth map and segmentation map using an off-the-shelf monocular depth estimator ZoeDepth~\cite{bhat2023zoedepth} and SAM~\cite{kirillov2023segment}, respectively. The depth map provides 3D spatial information, while the segmentation map delineates object boundaries, which we use together to approximate 3D bounding boxes of objects. For each segment, we perform back-projection, transforming the image segment using its depth map into a point cloud in 3D world space. We then estimate the corresponding oriented 3D bounding box with minimal volume for each back-projected segment point cloud and represent it as a cuboidal mesh. Finally, to obtain $D_c$, we render this scene of boxes using the original camera parameters used in back-projection and extract the depth map. This process is summarized in 
\cref{fig:pipeline-bbox-det}.

\begin{figure}[b!]
    \centering
    \includegraphics[width=\linewidth]{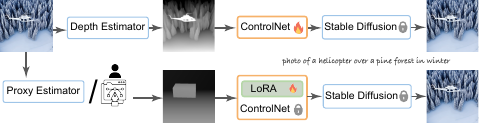}
    \caption{Training pipelines for ControlNet (\textit{Top}) and \name (\textit{Bottom}). Proxy Estimator represents our proxy depth extraction algorithms. During inference, we enable an option for a user to design the condition depth manually via a UI.}
    \label{fig:method-pipeline}
\end{figure}

\subsection{How to train for \name?}
We build our framework on StableDiffusion~\cite{rombach2022high} and ControlNet~\cite{zhang2023adding}.
In both scene boundary control and 3D box control, the constructed depth condition $D_c$ serves as a proxy for specifying a more generalized control - scene boundary and 3D bounding box, respectively - while retaining compatibility with ordinary depth-conditioned frameworks such as ControlNet. This ensures a smooth transition and facilitates the easy adoption of these approaches. Thanks to this backward compatibility, we can efficiently fine-tune a pre-trained depth ControlNet to achieve generalized depth control. We prepare the triplets (T,~$D_c$,~I) from the given (T,~I) pairs where $D_c$ is constructed using the algorithms presented in previous sections. We explore two primary options for fine-tuning: (i)~\textbf{Naive fine-tuning:} where we finetune the entire ControlNet; and (ii)~\textbf{LoRA fine-tuning:} where we employ LoRA-based fine-tuning of ControlNet (see \cref{fig:method-pipeline}). Specifically, for every attention block in ControlNet, we learn a low-rank update:
\begin{equation}
    W'\mathbf{x} = W\mathbf{x} + BA\mathbf{x}
\end{equation}
where $W \in \reals^{M\times N}$ is the original frozen projection, $B\in \reals^{M\times r}$ and $A\in \reals^{r \times N}$ are trainable low rank matrices of rank $r \ll min(M,N)$. It is important to note that in both cases Stable Diffusion U-net remains frozen during fine-tuning.
For both control types (C1 and C2), we use the NYU-Depth-v2~\cite{Silberman:ECCV12} dataset for fine-tuning and obtain textual image descriptions using BLIPv2~\cite{li2023blip} captioning.

\begin{figure}[b!]
    \centering
    \includegraphics[width=\linewidth]{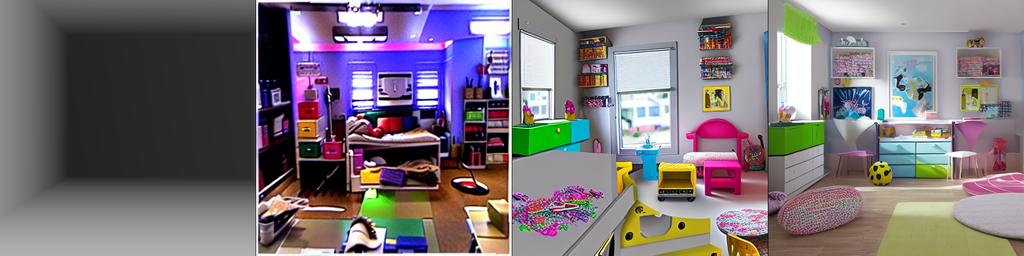}
    \caption{
    Given scene boundary~(left), comparing naive fine-tuning, adjusted skip weights, and our LoRA-based training, resp.}
    \label{fig:naive-skip-lora-comparison}
\end{figure}

\textbf{Inference time adjustments.}
We observe that naive fine-tuning results in saturated generations (\cref{fig:naive-skip-lora-comparison}). Interestingly, the color artifacts are eliminated if the residuals injected into Stable Diffusion U-net from ControlNet are reweighted at inference time. We observe that residual to the bottleneck block is responsible for most of the condition adherence and other higher resolution skip residuals largely add only local texture information. On the other hand, LoRA fine-tuning proves to be robust without any color issues. However, we introduce a controllable scale factor $\gamma$ at inference time such that the update is given as:
     $W'\mathbf{x} = W\mathbf{x} + \gamma BA\mathbf{x}$.
We observe that controlling $\gamma$ can lead to higher quality results, especially for $\gamma > 1$. We use $\gamma=1.2$ as the default.

\subsection{3D Box Editing}
Both the scene boundary condition and the box condition are easy to manipulate by a user. This opens up possibilities where a user can interactively manipulate the scene and objects, for example, by changing, adding, or removing the boxes. However, it is desirable to `lock in' and maintain the style and composition of the scene while manipulating the condition. A naive way to maintain the overall style would to be fix the seed but we observe that changing the condition even when fixing the seed can lead to diverging generations. To this end, we propose \textit{Style Preserving Edits}, inspired by video diffusion models~\cite{khachatryan2023text2video}, which allows users to maintain the desired style of the scene while conducting sequential edits. This feature is realized by replacing the ``keys" and ``values" for the given image in the attention layers of the Stable Diffusion U-net with those from the source image. However, in our case, sharing ``keys" and ``values" through all layers of the U-net leads to undesirable results, producing images identical to source images. We find that sharing ``keys" and ``values" from only the last two decoder layers ensures the preservation of the desired identity or style while adhering to the new target condition.

\begin{figure*}
\includegraphics[width=\linewidth]{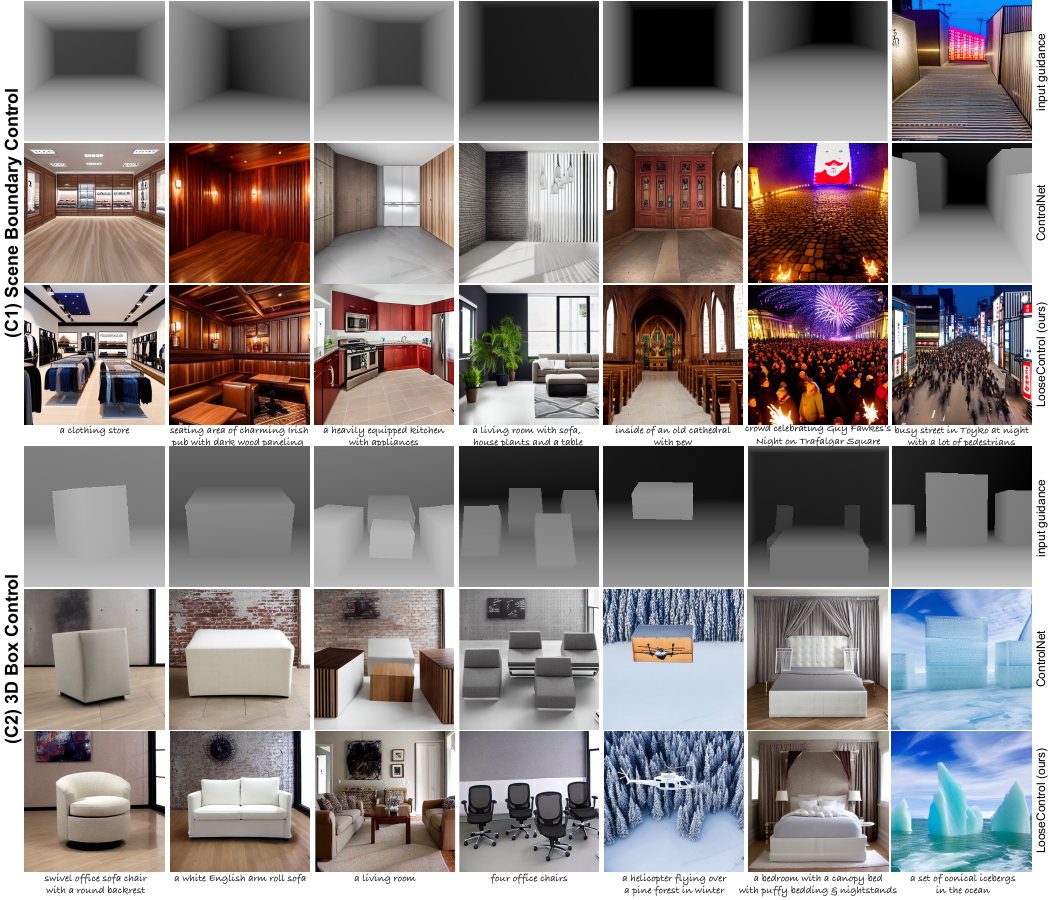}
        \caption{Qualitative results of (C1) Scene boundary control (rows 1-3) and (C2) 3D Box control (rows 4-6). \textit{Top}: Input depth condition \textit{Middle}: ControlNet generations. \textit{Bottom}: Ours. Our generations are more realistic and adhere to the prompt better than ControlNet.}
        \vspace{-.2in}
    \label{fig:layout-room-generation-comparison}
\end{figure*}

\subsection{Attribute Editing}
\name's generalized nature compared to ordinary depth control allows for a wider range of possible scenes under a given condition. To explore this expansive space, we conduct various forms of latent exploration.

ControlNet produces two types of residuals for a given condition: residuals added to the SD U-net bottleneck and residuals added to decoder skip connections. Our focus centers on the bottleneck feature space ($\mathcal{H}$-space), known for its semantic behavior~\cite{kwon2022diffusion,haas2023discovering} and from our experiments, its pivotal role in condition adherence.

In our exploration, we examine how the latent input, $\mathbf{x}$, affects ControlNet's output for a given fixed condition. We extract the most influential directions in x-space, which cause substantial changes in $\Delta \mathbf{h}\in \mathcal{H}$, and find their counterparts in $\mathcal{H}$-space, using Singular Value Decomposition (SVD) of the ControlNet Jacobian, $\partial \Delta \mathbf{h} /\partial \mathbf{x} $. To keep the SVD computation feasible, we only extract the first N directions, labeled $\{\mathbf{e}_i\}$, which guide our modification of the bottleneck residual via:
$\Delta \mathbf{h}' = \Delta \mathbf{h} + \beta \mathbf{e}_i$,
where the scalar $\beta$ controls the magnitude of the edit.

Per-condition latent exploration enhances user control by revealing various semantic edit directions, for example, sizes, types, and even the number of objects in the scene. This gives rise to a \textit{continuous} control space, allowing for fine-grained adjustments that were challenging with conventional conditioning methods. This continuous control space operates orthogonal to the input depth condition, enabling the maintenance of specified conditions—whether scene boundary control or 3D box control—while simultaneously exploring a wide range of generated images that adhere to the given condition.

\begin{figure*}
    \centering 
    \includegraphics[width=\linewidth]{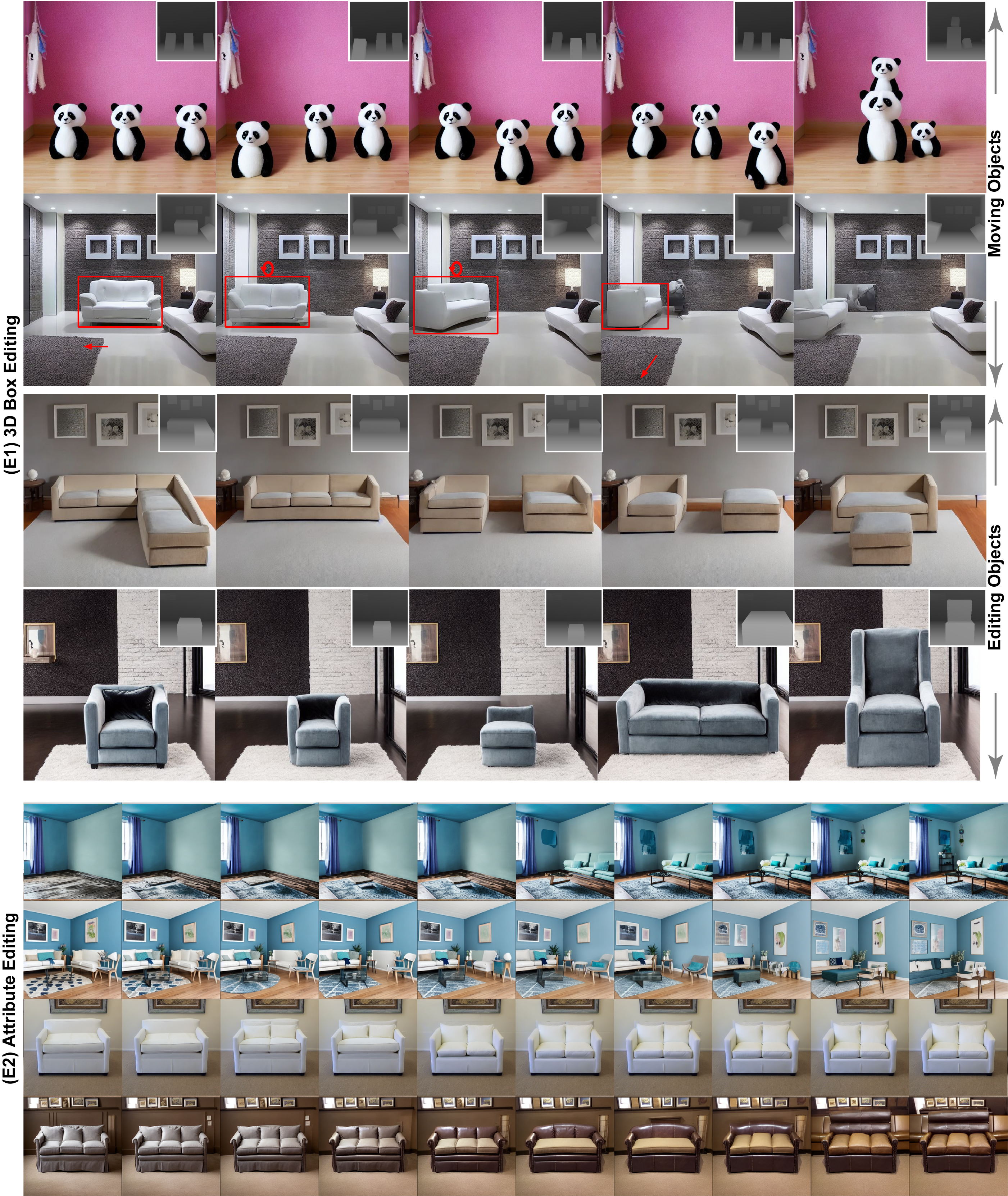}
    \caption{Qualitative results of (E1) 3D Box Editing, and (E2) Attribute Editing. (E1) 3D Box editing enables us to move and re-orient objects in the scene, change their size, split them, and more. (E2) Attribute Editing lets us change properties like the density of furniture (e.g., increase or decrease density) and the material of objects (e.g., fabric to leather or vica versa).}
    \label{fig:edit_plate}
\end{figure*}

\section{Applications and Results}
\label{sec:applications} 
We developed an interface to enable users to 
conveniently design a scene boundary and 3D boxes, and generate images interactively (see supplementary material). Here, we present qualitative and quantitative results.


\begin{figure}[b!]
    \centering
    \vspace{-.15in}
    \begin{overpic}[width=\linewidth,percent,tics=5]{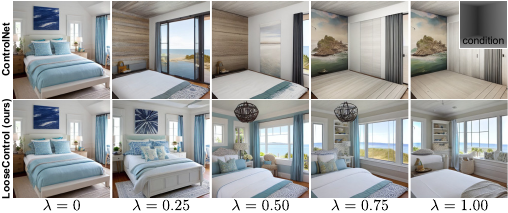}
    
    
    \end{overpic}
    \caption{Effect of conditioning control scale $\lambda$. \textit{Top}: ControlNet. \textit{Bottom}: Ours. The results were produced for the prompt \textit{``A cozy coastal bedroom with a sea view"}.}
    \label{fig:ablation-cond-scale}
\end{figure}


\paragraph{(C1) Scene Boundary Control.}
Scene boundary control enables diverse applications across various domains. We present some examples and comparisons between \name and ordinary control in ~\cref{fig:layout-room-generation-comparison}.
One particularly noteworthy application of scene boundary control is in indoor scene generation. Users can provide somewhat abstract specifications for room layouts, generally confined to walls and floor, and still generate images of fully furnished rooms. This is different from ordinary depth conditioning, which demands precise depth information.

Another scenario involves (partially) bounded outdoor scenes where the user can provide the scene boundary depth, for example, in terms of locations of buildings (See \cref{fig:teaser}). Unlike the ordinary depth control implemented by the original ControlNet, which leads to empty scenes, our \name model excels at generating realistic scenes, inclusive of objects like cars, traffic lights, and pedestrians, all while adhering to the user-provided scene boundary condition. This feat underscores the reliability and generalization capability of \name via proxy fine-tuning.

In \cref{fig:ablation-cond-scale}, we show the effect of varying the conditioning scale, $\lambda$, in the context of scene boundary control. We observe that ControlNet still produces `empty' rooms with lesser prompt adherence for most of the conditioning scales. 

\paragraph{(C2) 3D Box Control.}
We present generations and comparisons with ControlNet for 3D box control for a variety of scenes in \cref{fig:layout-room-generation-comparison}. ControlNet with ordinary control produces `boxy' generations which are not only unrealistic but also result in lesser prompt adherence. On the other hand, our method is able to produce more realistic images and shows better adherence to the given prompt demonstrating the utility of such control.

\paragraph{(E1) 3D Box Editing.}
We show qualitative results for 3D Box Editing in \cref{fig:edit_plate}. 3D box Editing allows users to manipulate objects, for example, by changing their position in 3D space, their orientation, and size. Users can perform 3D-aware edits, such as relocating furniture within a room or converting a small sofa chair to a big sofa. The \textit{Style Preserving Edit} mechanism facilitates sequential editing of object poses and locations, preserving the scene's essence while introducing creative modifications.
Since the 3D box control implicitly specifies the rough shape and size of the objects, we can edit the boxes themselves such as splitting a box into two, resulting in semantic edits on objects.

\noindent \textbf{(E2) Attribute Editing.}
Extracting the edit directions for a fixed scene boundary leads to continuous control directions for attributes such as the density of furniture and furniture type. 
For a given 3D box control condition (see \cref{fig:edit_plate}), the edit directions represent attributes like the shape, thickness, and material properties such as ``leather". 
We observe that edit directions are smooth and continuous for the most part and varying the magnitude $\beta$ directly reflects that magnitude of change in the attribute.

\paragraph{User Study.}
We conducted a user study to quantify the 
\begin{wrapfigure}[7]{r}{.58\columnwidth}
  \centering
  \vspace{-.1in}
    \includegraphics[width=.57\columnwidth]{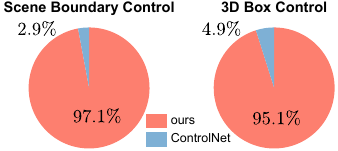} 
\end{wrapfigure}
improvement in control and quality of generations. 41 users were presented with the prompt and condition and asked to pick the image they preferred between ours and ControlNet. A majority of users, more than $95\%$, ranked our result better. See inset for results. 

\section{Conclusion}
We have presented \name to support 
generalized depth control for image generation. We introduced two types of control: (C1) Scene boundary control and (C2) 3D Box control to generate initial images. To refine initial results, we proposed (E1) 3D box editing and (E2) Attribute editing. Our framework provides new modes of creative generation and editing allowing users to more effectively explore the design space with depth guidance. A user study revealed over $95\%$ preference for \name compared to previous work.

\noindent \textbf{Limitations.}
Although our method works well for primary objects, control over secondary objects is harder to achieve -- we attribute this to the training data where secondary objects are less common. We expect a scene-based generator may lead to improved results. 
Also, similar to the original ControlNet we find that providing too many constraints as input reduces the diversity of the results. 

\noindent\textbf{Future work.}
We would like to use \name in conjunction with masks and also with MultiControlNet through the generalized specification of other maps. For example, in the context of sketch-based generation, specifying only a selection of (dominant) edges is easier for users than drawing all the Canny edges. 
Further, we would like to explore temporally consistent generation to produce output videos under smoothly changing guidance as in an interpolated 6DOF object specification.

{
    \small
    \bibliographystyle{ieeenat_fullname}
    \bibliography{main}
}

\clearpage
\setcounter{page}{1}

\twocolumn[{%
\renewcommand\twocolumn[1][]{#1}%
\maketitlesupplementary
\thispagestyle{empty}
\begin{center}
    \includegraphics[width=\linewidth]{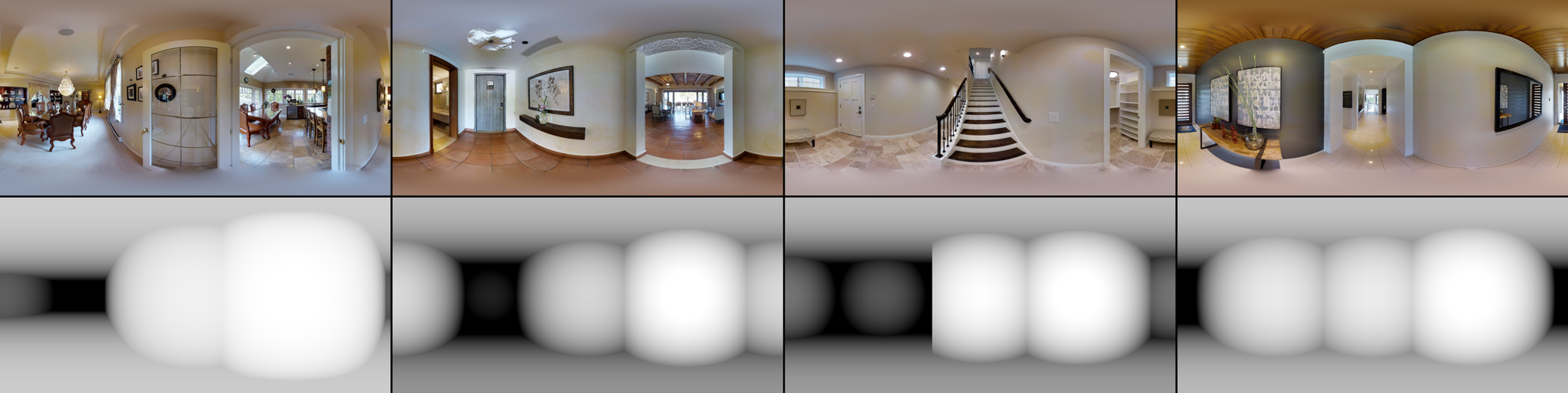}
        \captionof{figure}{\captionsize Upper-bound violations in Matterport3D~\cite{chang2017matterport3d}. \textit{Top:} RGB panorama images. \textit{Bottom:} Depth rendered using the layout labels from the dataset.} 
    \label{fig:room-violation}
\end{center}
}]


\section{Implementation details}
\label{sec:impl-details}
For all our experiments, we use Stable Diffusion~v1.5~\cite{rombach2022high} and the corresponding ControlNet~\cite{zhang2023adding} checkpoint \textit{``lllyasviel/control\_v11f1p\_sd15\_depth"} hosted on HugginFace~\cite{HuggingFace}. We use the PyTorch~\cite{paszke2019pytorch} framework and the diffusers~\cite{diffusers} library as the framework for diffusion models. We use ZoeDepth~\cite{bhat2023zoedepth} and SAM~\cite{kirillov2023segment} for extracting depth maps and segmentation maps respectively. For SAM, we use \verb|min_mask_area=1e4|. We make use of Pytorch3D~\cite{pyt3dravi2020accelerating} and Open3D~\cite{zhou2018open3d} in our 3D framework and PyTorch3D for rendering the depth maps for obtaining the proxy depths for fine-tuning. For backprojection, we randomly choose an FOV in the range of 43 and 57 degrees. For both Scene Boundary Control and 3D Box Control, we use LoRA rank $r=8$ and fine-tune only LoRA layers for $200$ steps with a learning rate of $0.0001$, and batch size of $12$ with Adam~\cite{kingma2014adam} optimizer. We use \verb|controlnet_conditioning_scale=1.0| and LoRA scale factor $\gamma=1.2$ as default. We built our 3D editor user interface using Gradio~\cite{abid2019gradio} and BabylonJS~\cite{babylonjs}.

\section{Issues with Room Layout Datasets}
As discussed in the main paper, a possible alternative for extracting the scene boundary for the preparation of the dataset for Scene Boundary Control could be room layout datasets. However, we note that these datasets often contain ambiguous layout labels that directly violate our tight upper-bound condition required to implement scene boundary control. We provide some examples of these violations for the popular MatterPort3D dataset as the representative in \cref{fig:room-violation}

\section{User study additional details}
We created an anonymous user study as a web form using streamlit~\cite{streamlit}. Users were presented with `Two-alternative forced choice' (2-AFC) with two options as images generated by our baseline (ControlNet) and our result and asked to respond with their preference for the given text prompt and condition image. The options were anonymized and did not indicate the name of the method. Each user was asked to respond to 10 randomized questions in total (5 for Scene Boundary Control, and 5 for 3D box control). The order of options was also randomized. As mentioned in the main paper, over 95\% of responses were in favor of our method.

%

\end{document}